\documentclass{article}
\usepackage{amssymb}
\usepackage[margin=1in]{geometry}
\usepackage{url}

\usepackage{listings}
\usepackage{xcolor}

\usepackage{subcaption} 

\usepackage{graphicx}
\usepackage{amsmath}         % KEEP - needed for equations
\usepackage{algorithm}
\usepackage{algpseudocode}

\usepackage{listings}
\usepackage{xcolor}
\usepackage{url}
\usepackage{hyperref}
\usepackage{booktabs}
\usepackage{caption}
\usepackage{subcaption}

\title{ISOPO: Proximal policy gradients without pi-old}

\author{Nilin Abrahamsen}
\date{}

\begin{document}

\maketitle

\begin{abstract}
This note introduces Isometric Policy Optimization (ISOPO), an efficient method to approximate the natural policy gradient in a single gradient step. In comparison, existing proximal policy methods such as GRPO or CISPO use multiple gradient steps with variants of importance ratio clipping to approximate a natural gradient step relative to a reference policy. In its simplest form, ISOPO normalizes the log-probability gradient of each sequence in the Fisher metric before contracting with the advantages. Another variant of ISOPO transforms the microbatch advantages based on the neural tangent kernel in each layer. ISOPO applies this transformation layer-wise in a single backward pass and can be implemented with negligible computational overhead compared to vanilla REINFORCE. 
\end{abstract}

\section{Introduction}

Reinforcement learning-based fine tuning has proven a powerful method to train reasoning and agentic behavior in LLMs. The GRPO algorithm and its variants (\cite{grpo,dr-grpo,dapo,opo,gspo}) have become the de facto standard for RL-finetuning of LLMs and are an adaptation of the classical Proximal Policy Optimization (PPO) \cite{ppo} algorithm with a simplified advantage estimator. 

Proximal policy methods such as GRPO can be viewed as efficient approximations to the \emph{natural policy gradient} (NPG) \cite{nat-pg} which itself would be impractical to implement as it involves the Fisher matrix whose size is quadratic in the number of parameters. Roughly, the NPG is defined to point towards the policy gradient but away from directions where the KL divergence grows the quickest. PPO/GRPO simulates an NPG step by taking multiple gradient steps with a surrogate objective based on a recent policy $\pi_{\text{old}}$ such that large changes in log-probabilities are not incentivized. This effectively constrains the updates to remain in a trust region near $\pi_{\text{old}}$ and stabilizes training. It is a simpler alternative to Trust-Region Policy Optimization (TRPO) \cite{trpo}, which uses a different surrogate with a hard constraint on the estimated KL divergence.

\begin{figure}[H]
    \centering
    \includegraphics[width=0.8\linewidth]{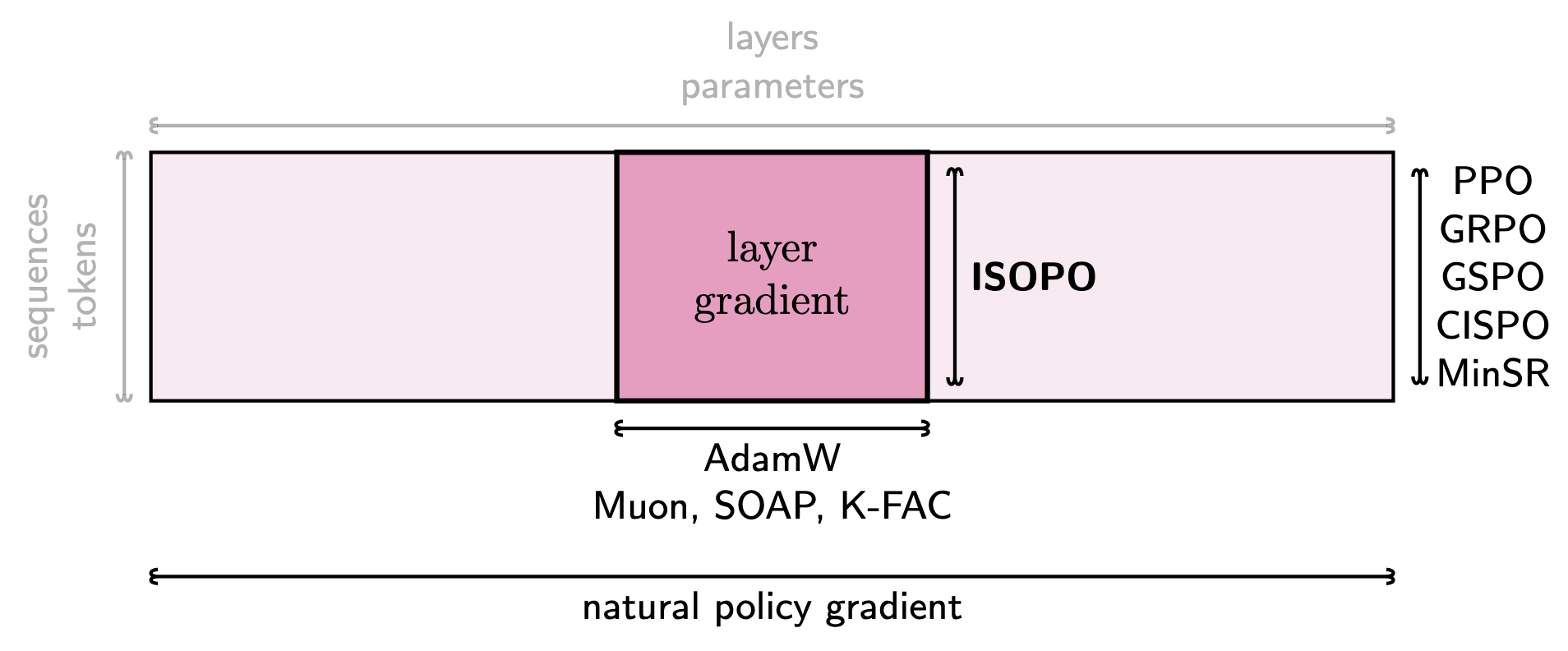}
    \caption{ISOPO acts on the batch dimension of each layer before the advantage vector is applied to the log-probability gradients. Right: GRPO and related methods acts on the same batch dimension, but by clipping sequences based on a reference policy instead of based on the gradient. Bottom: Many optimizers act on the parameter dimension, including diagonal multiplers (AdamW) and non-diagonal transformations (K-FAC, Shampoo, SOAP, Muon). The natural policy gradient itself is not tractable.}
    \label{fig:rectangle}
\end{figure}

\subsection{Related work}

Several methods to improve RL-based LLM training focus on different re-weighting or clipping schemes for the sample-wise (tokenwise or sequence-wise) objective \cite{wang20258020rulehighentropyminority,yang2025letlowprobabilitytokensoverdominate,cui2025entropymechanismreinforcementlearning}). 
CISPO \cite{CISPO} clips the sampling distribution to stabilize training without excluding tokens from the gradient calculation as in GRPO. Like these methods, ISOPO defines a re-weighting or transformation on the batch dimension (figure \ref{fig:rectangle} right), but unlike the aforementioned methods, ISOPO computes the re-weighting based on gradient information. Because the transformation in ISOPO happens in each layer individually, the ISOPO gradient can be computed in a single backward without introducing additional computational cost.

Natural gradient methods have been explored extensively in computational physics where the stochastic reconfiguration (SR) \cite{sr} method is analogous to the natural policy gradient, and the \emph{local energy} is analogous to the advantage function. The MinSR \cite{minsr} method improves efficiency of SR by preconditioning the batch dimension instead of the parameter dimension. Section \ref{sec:interacting} shows an interacting version of ISOPO which is analogous to a layer-wise version of the MinSR optimizer.

Another class of approximate natural gradient methods act on the parameter gradient vectors, typically layer-wise (figure \ref{fig:rectangle} bottom). These include the K-FAC \cite{kfac}, Shampoo \cite{shampoo}, and SOAP \cite{SOAP}. These methods pre-condition the rows and columns of a linear layer's weight matrix separately, an approximation that gives rise to the \emph{Kronecker-factored} (K-FAC) name. Another related optimizer is Muon \cite{jordan2024muon,muonclip} which unlike K-FAC operates only on the reduced gradients. Since these optimizers act on the parameter dimension, they can in principle be combined with ISOPO similarly to how AdamW and Muon optimizers are used with clipping-based GRPO. The experiments in this note use AdamW for easy comparison with the VeRL baselines \cite{verl}.

\subsection{Refresher on the natural policy gradient}

\label{sec:nat}
\newcommand\EE{\mathbb E}

Recall that the Fisher matrix $F$ is defined such that for a perturbation $v$ of the parameter vector $\theta$, the squared Fisher metric $\|v\|_F=v\cdot Fv$ is the squared perturbation of the log-probability, averaged over samples. That is,
\begin{equation}
\label{eq:vFv}
v\cdot Fv=\EE[|\nabla_\theta\log\pi_\theta(o|q)\cdot v|^2],
\end{equation}
where $q$ is the question and $o$ is the output of the model. This is to leading order the KL divergence incurred by the perturbation (or $\chi^2$ divergence, which is easier to see). The natural policy gradient is $v=F^{-1}g$ where $g=\EE[A(q,o)\nabla_\theta\log\pi_\theta(o|q)]$ is the vanilla policy gradient. Here $A(q,o)$ is the advantage of output $o$.
Importantly, the transformation $F^{-1}$ that defines the natural policy gradient does not involve the advantage $A(q,o)$. That is, $F^{-1}$ penalizes parameter directions with large policy moves, but it does not penalize large advantages.

\section{The non-interacting ISOPO gradient update}

The idea of the non-interacting ISOPO is to estimate the size of the gradient of each output sequence in the Fisher metric, i.e. how much each sequence contributes to the policy movement in terms of KL divergence. The sequences are then normalized in the Fisher metric before being multiplied by their advantages. This transformation is applied layer-wise during the backward pass (algorithm \ref{fig:algo-non-interacting}).

\begin{figure}[H]
    \centering
\begin{lstlisting}[language=Python]
# Compute non-interacting ISOPO microbatch gradient for a linear layer.
# Run backprop on objective = torch.sum(log_prob) without advantages.
def compute_ISOPO_grad_full_bwd_hook(mod, _, batch_grad_out):

    # Get samples to estimate Fisher norm sqrt(v * F v)
    overlap_samples = torch.randperm(microbatch_grad_out.shape[0])[:n_overlap_samples]
    a0 = act_in[overlap_samples]
    g0 = g_out[overlap_samples]

    # Reduce and re-scale sequence-wise
    grad = 0.0
    for seq_id, seq_indices in enumerate(microbatch.indices_by_sequence):
        act_in = mod.microbatch_act_in[seq_indices] # activations registered in forward pass
        grad_out = microbatch_grad_out[seq_indices]
        seq_logprob_grad = grad_out.T @ act_in

        F_norm = (
            torch.norm(torch.sum((g0 @ seq_logprob_grad) * a0, dim=1)) / 
            torch.norm(torch.norm(g0, dim=1) * torch.norm(a0, dim=1)) # optionally cache this
            )
        grad += microbatch.advantages[seq_id] * rescaling(seq_logprob_grad, F_norm=F_norm)
        
    # accumulate
    mod.grad += grad

def rescaling(grad, F_norm):
    # use e.g. EMA of trailing minibatch means for the regularization
    return 1 / (F_norm**2 + regularization).sqrt()

    

\end{lstlisting}
    \caption{The non-interacting variant of ISOPO scales the sequence gradients based on a stochastic estimate of the Fisher norm of the log-probability gradient.}
    \label{fig:algo-non-interacting}
\end{figure}

\subsubsection{Derivation: Sequence-wise bound on the KL contribution}
\label{sec:sample-wise}

The re-scaling is motivated as follows: Let $v=\nabla_\theta\log\pi_\theta(o|q)$ be the gradient of an output sequence's log-probability. The vanilla stochastic policy gradient of the sequence is $g = A(o|q) v$ where $A(o|q)$ is the advantage of the output. If the Fisher matrix were known, a stochastic natural gradient could be defined as $A(o|q) F^{-1}v$. 

Similarly to the stochastic natural policy gradient $A(o|q) F^{-1}v$, define the non-interacting ISOPO gradient as $A(o|q) w$ where $w$ is obtained from the log-probability gradient $v$. For the non-interaction ISOPO, define $w$ by re-scaling $v$ such that the KL divergence incurred by a perturbation $w$ is bounded by a constant.

By the discussion in section \ref{sec:nat}, the KL divergence incurred by update $w = \lambda v$ approximated by the squared Fisher metric $\lambda^2\|v\|_F^2$. So to bound the KL divergence by some constant $C^2$, it suffices to choose $\lambda\le C/\|v\|_F$. So define the layer-wise gradient update
\[
u_{l}=
\sum_{\text{output sequence }o}
\frac{A(o|q)}{\sqrt{\|\nabla_{\theta_l}\log\pi_\theta(o|q)\|_F^2+\text{reg}_l}}\nabla_{\theta_l}\log\pi_{\theta}(o|q),\]

where $l$ is the layer and $\text{reg}_l$ is some regularization. If the layer-wise regularization parameter is taken to be larger than the typical value of $\|\nabla_{\theta_l}\log\pi_\theta(o|q)\|_F^2$ then the transformation acts similarly to a clipping of the $\|\cdot\|_F$ norm.

\subsubsection{Estimating the Fisher metric}

To estimate the Fisher norm $\|v\|_F$, the direct way would be to use 

\begin{equation}\label{eq:F-norm-basic}\|v\|_F^2=v\cdot F v\approx v^T(\frac1n\sum_i g_ig_i^T)v=\frac1n\sum_i(v\cdot g_i)^2,\end{equation}

where $g_i$ are the reduced gradients corresponding to a batch of sequences. However, to increase the sample size with small computational and memory cost, ISOPO instead uses un-reduced token-position-wise\footnote{These are not exactly token-wise, as the activations at one position affect all tokens at subsequent positions.} gradients: Let $g$ be the reduction of $n$ token-position-wise gradients $g^{(j)}$ and use the approximation

\[
(v\cdot g)^2 = (v\cdot \sum_j g^{(j)})^2 \approx \sum_j(v\cdot g^{(j)})^2,\]

which heuristically assumes that the position-wise unreduced gradients are uncorrelated. To avoid bias due to the scale of the samples $g^{(j)}$, the estimate of $\|v\|_F^2$ will be normalized by the average scale of $\|g\|^2$ which is similarly estimated as $\|g\|^2=\|\sum_j g^{(j)}\|^2\approx\sum_j\|g^{(j)}\|^2$. This and \eqref{eq:F-norm-basic} yields the estimate

\begin{equation}
\label{eq:formal-F-norm}
\|v\|_F^2\approx \sum_j(v\cdot g^{(j)})^2/\sum_j\|g^{(j)}\|^2,
\end{equation}

where the sums are over position-wise unreduced gradients $g^{(j)}$. 

\newcommand\cdott{\boldsymbol{\cdot}}

\paragraph{Efficient implementation}
To calculate \eqref{eq:formal-F-norm}, note that in a linear layer, the weight update $v$ and the un-reduced gradient $g^{(j)}$ are matrices. Moreover, each matrix $g^{(j)}$ is a rank-one matrix, namely the outer product of the back-propagated $g_{\text{out},j}=\texttt{grad\_out[j]}$ and the input activation $a_{\text{in},j}=\texttt{act\_in[j]}$. Writing $v$ as $V$ to indicate its matrix shape and using $\cdott$ to denote the entry-wise dot product of matrices,

\begin{equation}
\label{eq:one}
\sum_j(v\cdot g^{(j)})^2 = \sum_j(V\cdott g_{\text{out}, j}a_{\text{in}, j}^T)^2
= \sum_j(g_{\text{out}, j}\cdot V a_{\text{in}, j})^2,
\end{equation}
\begin{equation}
\label{eq:two}
\sum_j\|g^{(j)}\|^2=
\sum_j(g_{\text{out}, j}a_{\text{in}}^T)\cdott(g_{\text{out}, j}a_{\text{in}}^T)
=
\sum_j(\|g_{\text{out}, j}\|\|a_{\text{in}}\|)^2
\end{equation}

Applying \eqref{eq:one} and \eqref{eq:two} to \eqref{eq:formal-F-norm} yields the estimate for $\texttt{F\_norm}$ in algorithm \eqref{fig:algo-non-interacting}:

\begin{equation}
\|v\|_F\approx{\sqrt{
\sum_j(g_{\text{out}, j}\cdot V a_{\text{in}, j})^2}}\:\Big/\:
{\sqrt{\sum_j(\|g_{\text{out}, j}\|\|a_{\text{in}}\|)^2}}.
\end{equation}

\subsubsection{Alternative sequence re-scalings}

\label{sec:alternative}

In section \ref{sec:sample-wise} the scaling was chosen to bound the KL divergence incurred by each sequence gradient uniformly. One consequence is that the update becomes bounded regardless of the scale of the stochastic policy gradient sample. This is in contrast with the formula $v=F^{-1}g$ for the natural gradient where v scales linearly with the stochastic policy gradient $g$, given a fixed estimate of the Fisher matrix. 

It is also possible to define the re-scaling based on the direction only without uniformly bounding the stochastic updates. For this, choose the update as the scalar multiple $\lambda v$ of the log-probability gradient $v=\nabla_{\theta_l}\log\pi_\theta(o|q)$ which is closest to the natural gradient estimate $F^{-1}g$ in the Fisher metric. That is, minimize
\begin{equation}\label{eq:to-be-min}
\|\lambda v-F^{-1}g\|_F^2=(\lambda v-F^{-1}g)^TF(\lambda v-F^{-1}g)
\end{equation}
over $\lambda$. This means solving for
\[0=v\cdot F(\lambda v-F^{-1}g)=\lambda v\cdot F v - v\cdot g.\]
Recall that $g=A(o|q)v$ is the vanilla stochastic policy gradient, so
$v\cdot g = A(o|q) v\cdot v$. So the minimizer of \eqref{eq:to-be-min} is $A(o|q)\|v\|^2/\|v\|_F^2$. This gives the alternative re-scaling,
\[
u_{l}=
\sum_{\text{output sequence }o}
{A(o|q)}\Big(\frac{\|\nabla_{\theta_l}\log\pi_\theta(o|q)\|_F^2}{\|\nabla_{\theta_l}\log\pi_\theta(o|q)\|^2}+\text{reg}_l\Big)^{-1}\nabla_{\theta_l}\log\pi_{\theta}(o|q).\]

Figure \ref{fig:algo-alternatives} shows a parameterized generalization of the ISOPO re-scaling which includes the examples of sections \ref{sec:sample-wise} and \ref{sec:alternative}.

%\begin{lstlisting}[basicstyle=\ttfamily\footnotesize\color{gray}]

\begin{figure}[H]
    \centering
%\begin{lstlisting}[language=Python,basicstyle=\ttfamily\footnotesize\color{gray}]
\begin{lstlisting}[language=Python,
    keywordstyle=\color{gray},
    commentstyle=\color{gray},
    stringstyle=\color{gray},
    numberstyle=\tiny\color{gray},
    frame=single
]
# alternative re-scalings
def rescaling(grad, F_norm, p=-1, q=0, r=0):
    rel_F_norm = F_norm / grad.norm()
    scaling = reg2(F_norm).pow(p) * reg2(grad.norm()).pow(q) * reg2(rel_F_norm).pow(r) 
    return scaling * grad

# regularize
def reg(val, regularization = 0.0):
    # use e.g. EMA of trailing minibatch means for the expectation
    return val + regularization * expectation(val) + 1e-8

# regularize the square of the value
def reg2(val, reg = 0.0):
    return reg(val**2, reg).sqrt()

\end{lstlisting}
    \caption{Generalized ISOPO re-scaling. The examples derived in sections \ref{sec:sample-wise} and \ref{sec:alternative} correspond to $p=-1$ ($q=r=0$) and $r=-2$ ($p=q=0$) respectively.}
    \label{fig:algo-alternatives}
\end{figure}

\begin{figure}[H]
    \centering
    \hspace*{-0.7cm}\includegraphics[width=0.7\linewidth]{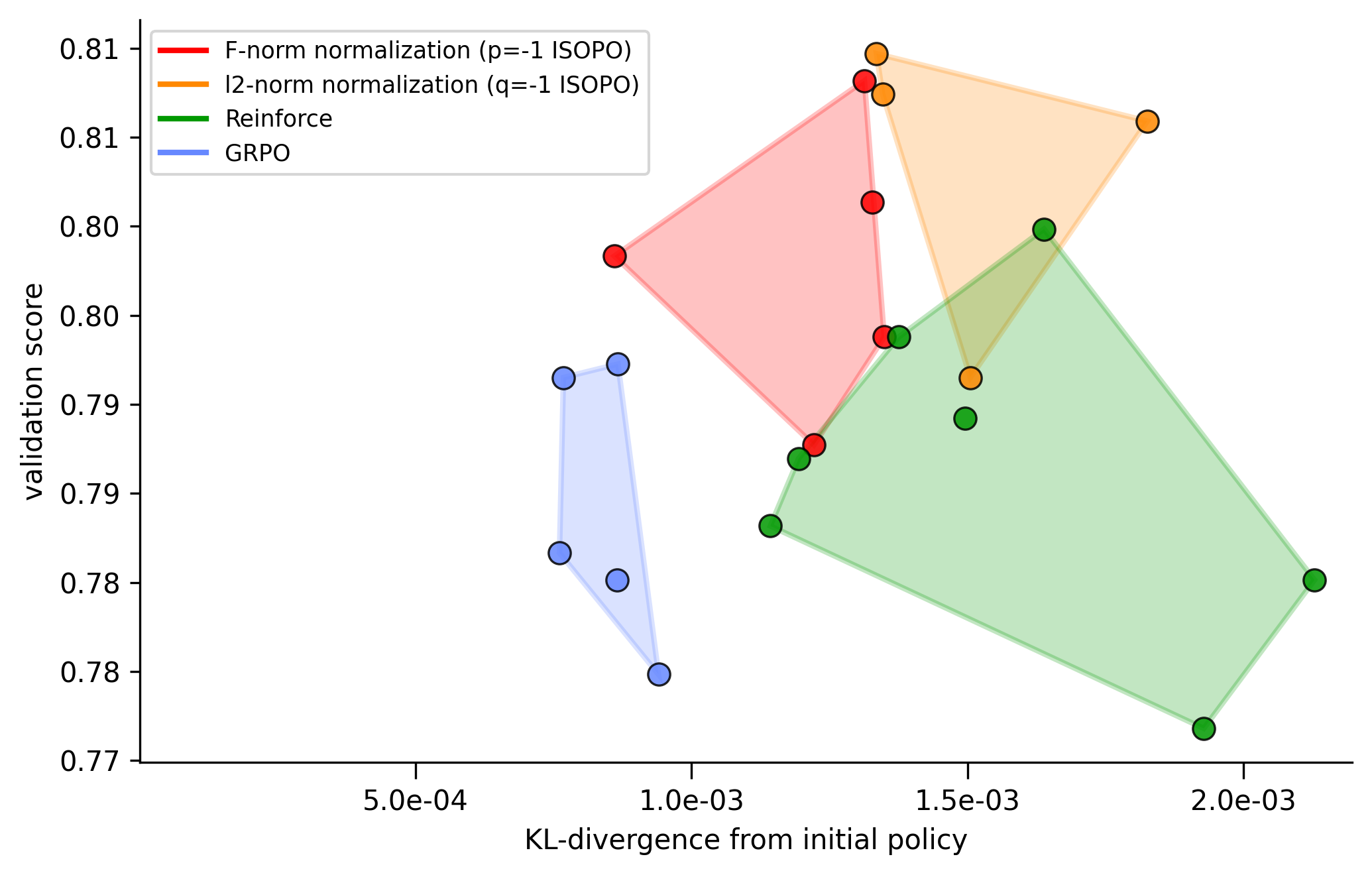}
    \caption{KL-drift and validation score at step 50 for non-interacting ISOPO vs GRPO and REINFORCE (see section \ref{sec:results} for the setup). Multiple runs are shown for each algorithm and the convex hulls are shaded for visual clarity. ISOPO is shown with two different settings: with Fisher normalization $p=-1$ described in section \ref{sec:sample-wise} (red) and with sequence-wise Euclidian normalization $q=-1$ (yellow). While both increase the validation score, only the $p=-1$ setting (red) decreases the KL drift.}
    \label{fig:kl-val}
\end{figure}

Figure \ref{fig:kl-val} shows KL divergence from the initial policy and validation score at step 50 for ISOPO with $p=-1$ (red) as described in section \ref{sec:sample-wise} vs sequence-wise euclidean normalization $q=-1$ (yellow). While both normalizations increase the validation score, only the Fisher normalization has the effect of decreasing the KL drift.

\section{Interacting variant}
\label{sec:interacting}

One can define a variant of ISOPO which includes interactions between the sequences in a microbatch. It is formally similar to a layer-wise application of the MinSR \cite{minsr} preconditioning from computational physics.

As in the re-scaling variant of ISOPO, during the backward pass the log-probability gradient of each sequence is computed for a given layer. In the interacting variant, stack these row vectors vertically to produce the Jacobian $J_{i,\cdot}=\nabla_{\theta_l}\log\pi_\theta(o_i|q)$ for the layer $l$. The empirical neural tangent kernel (NTK) $K=JJ^T$ for the layer has dimension $m\times m$ where $m$ is the number of sequences in the microbatch. The NTK is then used as a preconditioner by defining the update as
\[v=J^T(K+cI)^{-1}A,\]
where $A$ is the column vector of advantages, and $c$ is a Tikhonov regularization. This formula arises from choosing the vector in the span of the sampled sequence-wise log-probability gradients which maximizes the overlap with the vanilla policy gradient
$g=\frac1nJ^TA$
with a penalty on the Fisher metric $v\cdot Fv$ and the Euclidean norm (giving rise to the $c$). Here the Fisher metric within the subspace approximated by $\frac1n\|Jv\|^2$, which can be motivated by the definition \eqref{eq:vFv} of the Fisher. The interacting variant is shown in algorithm \ref{fig:algo-interacting}.

\begin{figure}[H]
    \centering
\begin{lstlisting}[language=Python]
# Compute interacting ISOPO microbatch gradient for a linear layer.
# Run backprop on objective = torch.sum(log_prob) without advantages.
def compute_ISOPO_grad_full_bwd_hook(mod, _, batch_grad_out):

    # Reduce sequence-wise
    seq_grads = []
    for seq_id, seq_indices in enumerate(microbatch.indices_by_sequence):
        act_in = mod.microbatch_act_in[seq_indices] # activations registered in forward pass
        grad_out = microbatch_grad_out[seq_indices]
        seq_grads.append( grad_out.T @ act_in )

    # make neural tangent kernel
    ntk = torch.empty((len(seq_grads), len(seq_grads)))
    for i in range(len(seq_grads)):
        for j in range(i, len(seq_grads)):
            ntk[i, j] = torch.sum(seq_grads[i] * seq_grads[j])
            ntk[j, i] = ntk[i, j]
            
    D, U = torch.linalg.eigh(ntk)
    preconditioner = 1.0 / (D + regularization)
    advantages_preconditioned = U @ (preconditioner * (U.T @ microbatch.advantages))

    mod.grad += torch.stack(seq_grads, dim=-1) @ advantages_preconditioned 
        
\end{lstlisting}
    \caption{The interacting variant of ISOPO does not use token-position-wise gradients like the non-interaction variant. Instead it uses the reduced gradient of each sequence to construct the empirical neural tangent kernel.}
    \label{fig:algo-interacting}
\end{figure}

\section{Results}

\label{sec:results}

\begin{figure}[H]
    \centering
    \includegraphics[width=0.7\linewidth]{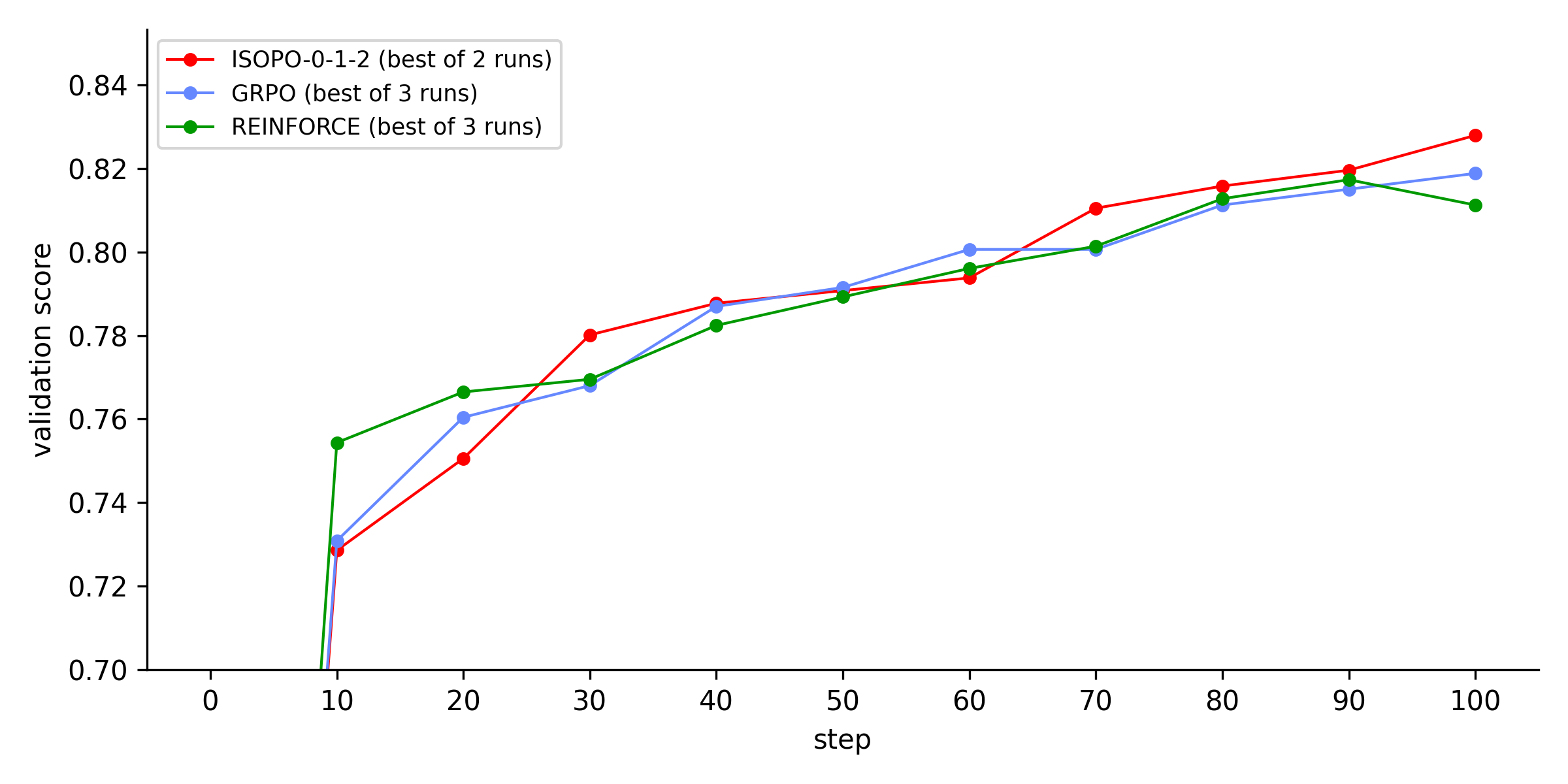}
    \caption{Validation curve for non-interacting ISOPO (red) with $p=0$, $q=-1$ and $r=-2$ (See algorithm \ref{fig:algo-alternatives}) and no regularization. Baselines are REINFORCE (green) with no clipping and GRPO (blue) which uses PPO clipping. For every 10 steps the best validation score among 3 runs is shown for each algorithm (except only best of 2 for ISOPO).}
    \label{fig:val-non-interacting}
\end{figure}

I implemented ISOPO in VeRL \cite{verl}. The experiments shown are for gsm8k with the Qwen-3 0.6B model. All runs use the group-relative advantage estimator, and all runs are without KL penalty. The two baselines shown are GRPO and REINFORCE. Here, GRPO refers to using PPO clipping of the importance sampling ratio, and REINFORCE refers to the same algorithm with no clipping. Like REINFORCE, ISOPO is run without PPO clipping. Figure \ref{fig:val-non-interacting} shows a comparison between best-of-3 validation curves for REINFORCE (green), GRPO (blue), and ISOPO (red) with $q=-1$ and $r=-2$. The code for ISOPO will be available at \url{https://github.com/nilin/isopo}.

% and have \texttt{train\_batch\_size = 128 = 2*ppo\_mini\_batch\_size}.

\begin{figure}[H]
    \centering
    \includegraphics[width=0.8\linewidth]{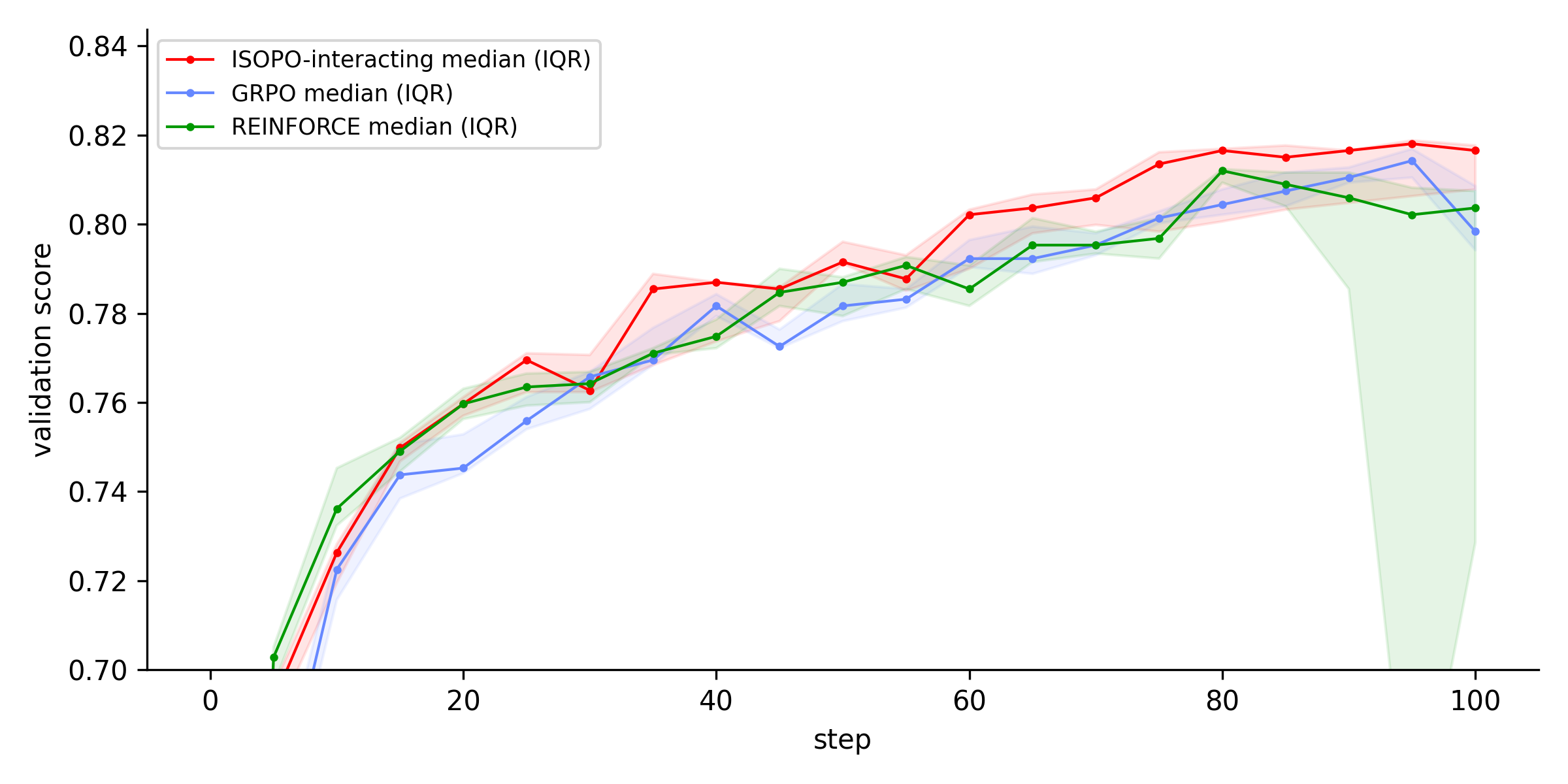}
    \caption{Validation curve for interacting ISOPO with regularization factor $1.0*\EE[\text{mean}(D)]$ which is computed as an average over previous minibatches (EMA over minibatches with 0.9 decay). For every 5 steps the median validation score among 3 runs is shown, along with the range of validation scores (shaded area).}
    \label{fig:val-interacting}
\end{figure}

\bibliographystyle{abbrv}
\bibliography{bibliography}
\end{document}